\documentclass{article}
\usepackage[utf8]{inputenc}

\usepackage[colorlinks=true, linkcolor=blue, citecolor=blue, urlcolor=blue]{hyperref}

\usepackage{setspace}
\usepackage{multicol}
\usepackage[dvipsnames]{xcolor}

\usepackage{makecell}
\usepackage{tabularx}

\title{\textbf{mwBTFreddy: A Dataset for Flash Flood Damage Assessment in Urban Malawi}}
\usepackage{authblk}
\makeatletter
\let\AB@footnotemark\@footnotemark
\let\AB@footnotetext\@footnotetext
\makeatother

\author[1]{Evelyn Chapuma}
\author[1]{Grey Mengezi}
\author[2]{Lewis Msasa}
\author[1]{Amelia Taylor\thanks{Corresponding author: admin@kailab.tech}}
\affil[1]{Kuyesera AI Lab, Malawi University of Business and Applied Sciences}
\affil[2]{University of California, Berkeley}
\date{April 2025}

\usepackage[a4paper, margin=1in]{geometry}
\usepackage{cite}
\usepackage{url}
\begin{document}

\maketitle

\begin{abstract}
\textit{This paper describes the mwBTFreddy dataset, a resource developed to support flash flood damage assessment in urban Malawi, specifically focusing on the impacts of Cyclone Freddy in 2023. The dataset comprises paired pre- and post-disaster satellite images sourced from Google Earth Pro, accompanied by JSON files containing labelled building annotations with geographic coordinates and damage levels (no damage, minor, major, or destroyed). Developed by the Kuyesera AI Lab at the Malawi University of Business and Applied Sciences, this dataset is intended to facilitate the development of machine learning models tailored to building detection and damage classification in African urban contexts. It also supports flood damage visualisation and spatial analysis to inform decisions on relocation, infrastructure planning, and emergency response in climate-vulnerable regions.}
\end{abstract}

\section{Introduction}

Flash floods are among the most destructive natural disasters, often resulting in substantial economic losses and posing severe threats to human life and infrastructure. Their sudden onset makes accurate information and rapid response critical for minimising damage. Satellite imagery provides an invaluable means of remotely assessing flood impact and holds significant promise for automated damage detection and disaster risk reduction\cite{Gupta, Ma}.
Urban areas are particularly susceptible to flash flooding due to the prevalence of impervious surfaces, informal settlements in risk areas and inadequate drainage systems. In Malawi, a study conducted in Mzuzu city demonstrated the utility of geospatial and hydrological modelling to map urban flood hazards, underlining the urgent need for data-driven approaches to risk reduction\cite{Gumindoga}. However, the availability of data for assessing flash flood damage, particularly in informal urban settlements located on mountainous slopes, remains limited.
To address this gap, we developed a dataset comprising pre- and post-Cyclone Freddy satellite images covering flood-affected urban areas in Blantyre, Malawi. This dataset aims to support the evaluation of machine learning approaches for automatic damage classification, with a particular focus on flash flood impact in densely populated and ecologically vulnerable regions.
\subsection{Context}

Blantyre, Malawi’s main commercial hub, has witnessed rapid population growth and urban sprawl. Along with Lilongwe, Blantyre contributes approximately 31\% to Malawi’s GDP. Recent studies show that built-up areas in the city have expanded at an average annual rate of 7\% over the past decade\cite{mawenda2020urban}. The rise of informal settlements—particularly on and around mountainous terrain—has placed growing pressure on the region's green spaces. Housing shortages have driven communities to settle in ecologically fragile areas, including the three mountains in Blantyre, leading to widespread deforestation and environmental degradation.

Soche Mountain, once covered by over 150 hectares of rainforest, has been especially affected. Settlements, subsistence farming, and other human activities have led to extensive removal of vegetation. The loss of this natural cover has severely reduced the mountain’s ability to absorb rainfall, thereby increasing surface runoff and the risk of flash floods and mudslides. These changes pose serious threats to communities living along the mountain's slopes. Ndirande Mountain, another key ecological landmark in Blantyre has experienced probably an even worse ecological deterioration and it sits largely bare of forests. Uncontrolled settlements have taken over the slopes of the mountain.

In March 2023, Tropical Cyclone Freddy struck Southern Africa, affecting Madagascar, Mozambique, and Malawi. In Malawi, the cyclone caused widespread flooding, fatalities, and major disruptions to services such as healthcare and education\cite{lubanga2023freddy}. In Blantyre, the areas surrounding the Soche and Ndirande Mountains experienced some of the most severe impacts, with flash floods and mudslides devastating local communities\cite{kadzuwa2023freddy}. The environmental degradation on the mountain amplified the effects of the cyclone, making clear the connection between urban planning, ecological preservation, and risk of disasters\cite{nazombe2023greenspace}.

The dataset presented here offers a foundation for analysing flash flood and mudslide damage through satellite imagery. Such analyses can be enhanced by incorporating community-generated data and modelling future climate scenarios. The ultimate goal is to support better-informed urban planning decisions, including the development of effective drainage systems, strategic vegetation management, and safer land use practices in risk-prone areas.

\section{Datasheet}

\definecolor{darkblue}{RGB}{46,25, 110}

\newcommand{\dssectionheader}[1]{%
   \noindent\framebox[\columnwidth]{%
      {\fontfamily{phv}\selectfont \textbf{\textcolor{darkblue}{#1}}}
   }
}

\newcommand{\dsquestion}[1]{%
    {\noindent \fontfamily{phv}\selectfont \textcolor{darkblue}{\textbf{#1}}}
}

\newcommand{\dsquestionex}[2]{%
    {\noindent \fontfamily{phv}\selectfont \textcolor{darkblue}{\textbf{#1} #2}}
}

\newcommand{\dsanswer}[1]{%
   {\noindent #1 \medskip}
}

\begin{singlespace}

\begin{multicols}{2}

\dssectionheader{Motivation}

\dsquestionex{For what purpose was the dataset created?}{Was there a specific task in mind? Was there a specific gap that needed to be filled? Please provide a description.}

\dsanswer{This dataset was developed to support the assessment of damage to buildings and infrastructure caused by flash floods in Blantyre, Malawi. It addresses a critical gap in localised disaster response tools by providing high-resolution satellite imagery specific to the region’s geographical and architectural context.

The dataset comprises pre- and post-disaster satellite images from three urban areas in Blantyre—Chilobwe, Ndirande, and Chirimba—which were among the most severely affected during Cyclone Freddy. Blantyre was the only major urban centre in Malawi to experience such extensive and concentrated damage. However, the cyclone had a broad and uneven impact, triggering significant flash flooding and landslides in rural and peri-urban communities in other parts of the country. However, our focus was on damage in urban areas, hence our focus on Blantyre.

This data set enables key tasks such as building detection, structural damage classification, and impact analysis of infrastructure. As buildings in Malawi typically reflect region-specific construction styles and materials, machine learning models trained on imagery from other countries may not generalise well to the Malawian context. A localised dataset such as this is therefore needed for developing accurate and context-aware models to support disaster risk reduction, emergency response, and resilient urban planning.
}
\end{multicols}

\begin{center}
\textbf{Table I. Image Counts and Coordinates of the Selected Areas} \\[0.5em]
\begin{tabular}{|c|c|c|}
\hline
Area & Coordinates & No. Images \\
\hline
Chirimba & \makecell[l]{35.06135646185926,-15.7455674468705\\35.0630140330653,-15.74401809140623\\35.0606258243407,-15.74217357016835\\35.05878123267473,-15.74352129840774
} & 6 \\
\hline
Ndirande & \makecell[l]{35.04990122048631,-15.76751173684101\\35.04753585118715,-15.76823081389567\\35.05147913477497,-15.7744571861299\\35.05471668399898,-15.77301319505322
} & 20 \\
\hline
Chilobwe & \makecell[l]{35.00004428168152,-15.82503923829606\\35.04009739154355,-15.82507457317074\\35.04012666318832, -15.85010734237624\\35.00015291121928, -15.85003454786558
} & 1000 \\
\hline
Total & \makecell[l]{
} & 1026 \\
\hline
\end{tabular}
\end{center}

\begin{multicols}{2}
\dsquestion{Who created this dataset (e.g., which team, research group) and on behalf of which entity (e.g., company, institution, organization)?}
\dsanswer{The dataset was created by a team of researchers working on the Disaster Management (DIMA) project under the Kuyesera AI Lab at MUBAS.

\begin{center}
\textbf{Table II. The Team} \\[0.5em]
\begin{tabular}{|c|c|c|}
\hline
ID & Name & Role \\
\hline
1 & \makecell[l]{Dr. Amelia Taylor
} & Project Manager \\
\hline
2 & \makecell[l]{Evelyn Chapuma
} & Dataset Creator \\
\hline
3 & \makecell[l]{Grey Mengezi
} & Dataset Creator \\
\hline
4 & \makecell[l]{Lewis Msasa
} & Research Associate \\
\hline

\hline
\end{tabular}
\end{center}
}

\dsquestionex{Who funded the creation of the dataset?}{If there is an associated grant, please provide the name of the grantor and the grant name and number.}

\dsanswer{The research did not receive any dedicated funding.}

\dsquestion{Any other comments?}

\dsanswer{No.}

\bigskip
\dssectionheader{Composition}

\dsquestionex{What do the instances that comprise the dataset represent (e.g., documents, photos, people, countries)?}{ Are there multiple types of instances (e.g., movies, users, and ratings; people and interactions between them; nodes and edges)? Please provide a description.}

\dsanswer{Each instance in the dataset comprises either a pre- or post-disaster satellite image accompanied by a corresponding JSON annotation file. The annotation file includes the geographic coordinates of the image, polygonal outlines of building footprints, associated pixel values, and a damage classification for each identified structure.
Cyclone Freddy affected Malawi from the beginning od March 2023, with the first flash floods occurring on 11th March. Images were captured before and after Cyclone Freddy, with pre-disaster images from August, 2022, and post-disaster images from May, 2023, at 1024x768 resolution and 70m scale. These dates were chosen based on the availability of the data on Google Earth Pro.
}

\dsquestion{How many instances are there in total (of each type, if appropriate)?}

\dsanswer{There are a total of 696 anotated instances: 348 instances for pre-disaster images and 348 instances of post-disaster images.

\begin{center}
\label{tb:instances}
\textbf{Table III. Instances of the Dataset} \\[0.5em]
\begin{tabular}{|c|c|c|c|}
\hline
Entity & Pre & Post & Total \\
\hline
Images & \makecell[l]{348
} & 348 & 696 \\
\hline
JSON files & \makecell[l]{348
} & 348 & 696 \\
\hline
\end{tabular}
\end{center}
}

\dsquestionex{Does the dataset contain all possible instances or is it a sample (not necessarily random) of instances from a larger set?}{ If the dataset is a sample, then what is the larger set? Is the sample representative of the larger set (e.g., geographic coverage)? If so, please describe how this representativeness was validated/verified. If it is not representative of the larger set, please describe why not (e.g., to cover a more diverse range of instances, because instances were withheld or unavailable).}

\dsanswer{The dataset contains all possible instances.}

\dsquestionex{What data does each instance consist of? “Raw” data (e.g., unprocessed text or images) or features?}{In either case, please provide a description.}

\dsanswer{Each instance is an image (pre or post) with its corresponding JSON file. The images are raw TIFF files that have been geo-tagged. The JSON file has coordinates (both pixel and geographic coordinates) of the different building polygons and their damage classification status (no-damage, minor-damage, major-damage, destroyed).
}

\dsquestionex{Is there a label or target associated with each instance?}{If so, please provide a description.}

\dsanswer{A consistent file naming convention is used throughout the dataset. Pre- and post-disaster images corresponding to the same geographical area share the same base name and are distinguished by the affix "pre" or "post". Each image has a corresponding JSON annotation file with an identical name (including the affix), ensuring a clear one-to-one relationship between the image and its metadata.
}

\dsquestionex{Is any information missing from individual instances?}{If so, please provide a description, explaining why this information is missing (e.g., because it was unavailable). This does not include intentionally removed information, but might include, e.g., redacted text.}

\dsanswer{Yes. The following buildings were excluded from the dataset instances:
\begin{enumerate}
    \item Buildings that were not clearly visible or could not be clearly demarcated.
    \item Buildings that were 20\% obscured by trees.
    \item Buildings whose larger or significant portions appeared in a different image.
    \item We relied on the availability of the data around Cyclone Freddy from Google Earth Pro, hence the post and pre images may not be a faithfull reflection of the damage.
\end{enumerate}
}

\dsquestionex{Are relationships between individual instances made explicit (e.g., users’ movie ratings, social network links)?}{If so, please describe how these relationships are made explicit.}

\dsanswer{The relationship between pre- and post-disaster images and their corresponding JSON annotation files is established through a consistent and descriptive file naming convention.
}

\dsquestionex{Are there recommended data splits (e.g., training, development/validation, testing)?}{If so, please provide a description of these splits, explaining the rationale behind them.}

\dsanswer{No.}

\dsquestionex{Are there any errors, sources of noise, or redundancies in the dataset?}{If so, please provide a description.}

\dsanswer{The following could be the sources of noise. 
\begin{enumerate}
    \item 
    The resolution of the satellite images used was not sufficient to clearly identify some buildings. 
    \item Some buildings were partially or completely obscured by trees which affected the accuracy of the building polygons.
    \item Some buildings had a large or significant parts across several images.
    \item Geographical coordinates generated through georeferencing may not be 100\% accurate.
\end{enumerate}
}

\dsquestionex{Is the dataset self-contained, or does it link to or otherwise rely on external resources (e.g., websites, tweets, other datasets)?}{If it links to or relies on external resources, a) are there guarantees that they will exist, and remain constant, over time; b) are there official archival versions of the complete dataset (i.e., including the external resources as they existed at the time the dataset was created); c) are there any restrictions (e.g., licenses, fees) associated with any of the external resources that might apply to a future user? Please provide descriptions of all external resources and any restrictions associated with them, as well as links or other access points, as appropriate.}

\dsanswer{The dataset is self-contained. 
}

\dsquestionex{Does the dataset contain data that might be considered confidential (e.g., data that is protected by legal privilege or by doctor-patient confidentiality, data that includes the content of individuals non-public communications)?}{If so, please provide a description.}

\dsanswer{N/A.
}

\dsquestionex{Does the dataset contain data that, if viewed directly, might be offensive, insulting, threatening, or might otherwise cause anxiety?}{If so, please describe why.}

\dsanswer{No
}

\dsquestionex{Does the dataset relate to people?}{If not, you may skip the remaining questions in this section.}

\dsanswer{The view which was used to capture the images on google earth pro hid all labels such as names of streets, buildings etc.
}

\dsquestionex{Does the dataset identify any subpopulations (e.g., by age, gender)?}{If so, please describe how these subpopulations are identified and provide a description of their respective distributions within the dataset.}

\dsanswer{No
}

\dsquestionex{Is it possible to identify individuals (i.e., one or more natural persons), either directly or indirectly (i.e., in combination with other data) from the dataset?}{If so, please describe how.}

\dsanswer{No
}

\dsquestionex{Does the dataset contain data that might be considered sensitive in any way (e.g., data that reveals racial or ethnic origins, sexual orientations, religious beliefs, political opinions or union memberships, or locations; financial or health data; biometric or genetic data; forms of government identification, such as social security numbers; criminal history)?}{If so, please provide a description.}

\dsanswer{No
}

\dsquestion{Any other comments?}

\dsanswer{No
}

\bigskip
\dssectionheader{Collection Process}

\dsquestionex{How was the data associated with each instance acquired?}{Was the data directly observable (e.g., raw text, movie ratings), reported by subjects (e.g., survey responses), or indirectly inferred/derived from other data (e.g., part-of-speech tags, model-based guesses for age or language)? If data was reported by subjects or indirectly inferred/derived from other data, was the data validated/verified? If so, please describe how.}

\dsanswer{Satellite imagery of the selected areas was exported directly from Google Earth Pro, using pre- and post-Cyclone Freddy dates chosen to be as close as possible to the flash flood event. Care was taken to ensure that the selected images were cloud-free and of high visual quality. To systematically capture the imagery, gridlines were overlaid using a custom script that generated grids with specific latitude and longitude spacing, ensuring that image tiles were contiguous but non-overlapping. Each image was labelled according to its corresponding grid cell identifier and georeferenced in QGIS using the four corner coordinates of the grid cell.

A Python script was then used to crop and resize the images, removing irrelevant borders and content, and to apply a consistent naming convention. Annotation was conducted in QGIS following the xBD methodology, which involves classifying building damage based on visual assessment. The annotations were initially stored in CSV format and subsequently converted into JSON files using a custom script. These JSON files contain building polygons, geographic coordinates, pixel values, and damage classification levels. The result is a complete dataset of cropped, geo-referenced, and annotated satellite images, suitable for training and evaluating machine learning models in disaster impact assessment.
}

\dsquestionex{What mechanisms or procedures were used to collect the data (e.g., hardware apparatus or sensor, manual human curation, software program, software API)?}{How were these mechanisms or procedures validated?}

\dsanswer{The labelling process involved complete geotagging of images by assigning geographic coordinates to all features, ensuring that each building polygon is accurately referenced in space. Georeferencing was done using four corner points of each grid cell—representing the boundaries of the image tile. These coordinates were used to align the imagery spatially within a GIS environment.
We evaluated three tools for georeferencing and annotation and ultimately selected QGIS due to its open-source nature, strong geospatial capabilities, and compatibility with the xBD damage classification methodology. Polygons representing buildings were manually drawn in QGIS and saved as shapefiles. Each polygon was labelled as a feature and classified using the xBD Joint Damage Scale, which categorises damage into four classes: no-damage, minor-damage, major-damage, and destroyed. The pre-disaster images for each tile was annotated before their corresponding post- images. This process helped us ensure consistency in building identification and damage assessment. Buildings that were not clearly visible or distinguishable in either the pre- or post-disaster image were also excluded from annotation. The exclusion criteria for buildings is as follows.
\begin{itemize}
    \item Buildings that were present in the post-image but were not present in the pre-image, e.g. buildings that were constructed after the disaster or were not complete pre-disaster
    \item Buildings that were heavily occluded by vegetation were ignored. A building was ignored if it was at least 20\% occluded.
    \item Buildings that were not clearly demarcated because of resolution.
    \item Those buildings that appeared in more than two images because they were cut at the grids were only annotated in the image where they were dominant.
\end{itemize}
A script was used to generate CSV files from the labels(shapefiles) which were then used to generate JSON files corresponding to each image using another script. 

\begin{flushright}
\textbf{Table IV. xBD Classification Table} \\[0.5em]
\begin{tabular}{|l|p{4cm}|}
\hline
\textbf{Disaster Level} & \textbf{Structure Description} \\
\hline
0 (No Damage) & Undisturbed. No sign of water, structural or shingle damage. \\
\hline
1 (Minor Damage) & Building partially damaged, water surrounding structure, roof elements missing, visible large cracks. \\
\hline
2 (Major Damage) & Partial wall or roof collapse, or surrounded by water/mud. \\
\hline
3 (Destroyed) & Completely collapsed, partially/completely covered with water/mud, or otherwise no longer present. \\
\hline
\end{tabular}
\end{flushright}

From the 1026 images, we did not annotate those that contained no buildings, for example images that covered the mountain terrains and fields.

\begin{flushleft}
\textbf{Table V. Image Selection Summary Table} \\[0.5em]
\begin{tabular}{|l|c|c|}
\hline
\textbf{Category} & \textbf{Count} & \textbf{Percentage} \\
\hline
Total Images & 1026 & 100\% \\
Unannotated Images & 330 & 32\% \\
Annotated Images & 696 & 68\% \\
\hline
\end{tabular}
\end{flushleft}

\dsquestion{If the dataset is a sample from a larger set, what was the sampling strategy (e.g., deterministic, probabilistic with specific sampling probabilities)?}

\dsanswer{N/A.}

\dsquestion{Who was involved in the data collection process (e.g., students, crowdworkers, contractors) and how were they compensated (e.g., how much were crowdworkers paid)?}

\dsanswer{No primary data collection was conducted. The dataset was developed using publicly available satellite imagery exported from Google Earth Pro. The data processing and annotation were carried out by members of the Kuyesera AI Lab, along with an intern who was a student at the University of California, Berkeley. The intern participated as part of an institutional internship programme supported by the university. No external contractors or paid crowdworkers were involved in the process.}

\dsquestionex{Over what timeframe was the data collected? Does this timeframe match the creation timeframe of the data associated with the instances (e.g., recent crawl of old news articles)?}{If not, please describe the timeframe in which the data associated with the instances was created.}

\dsanswer{The development of the dataset and pipeline for creating the geotagging and annotation took four months.}

\dsquestionex{Were any ethical review processes conducted (e.g., by an institutional review board)?}{If so, please provide a description of these review processes, including the outcomes, as well as a link or other access point to any supporting documentation.}

\dsanswer{N/A.}

\dsquestionex{Does the dataset relate to people?}{If not, you may skip the remaining questions in this section.}

\dsanswer{N/A.}

\dsquestion{Did you collect the data from the individuals in question directly, or obtain it via third parties or other sources (e.g., websites)?}

\dsanswer{N/A.}

\dsquestionex{Were the individuals in question notified about the data collection?}{If so, please describe (or show with screenshots or other information) how notice was provided, and provide a link or other access point to, or otherwise reproduce, the exact language of the notification itself.}

\dsanswer{N/A.}

\dsquestionex{Did the individuals in question consent to the collection and use of their data?}{If so, please describe (or show with screenshots or other information) how consent was requested and provided, and provide a link or other access point to, or otherwise reproduce, the exact language to which the individuals consented.}

\dsanswer{N/A.}

\dsquestionex{If consent was obtained, were the consenting individuals provided with a mechanism to revoke their consent in the future or for certain uses?}{If so, please provide a description, as well as a link or other access point to the mechanism (if appropriate).}

\dsanswer{N/A.}

\dsquestionex{Has an analysis of the potential impact of the dataset and its use on data subjects (e.g., a data protection impact analysis) been conducted?}{If so, please provide a description of this analysis, including the outcomes, as well as a link or other access point to any supporting documentation.}

\dsanswer{N/A.}

\dsquestion{Any other comments?}

\dsanswer{We initially collected 1,026 satellite images covering the geographical area affected by flash floods around Soche and Ndirande mountains in Blantyre, Malawi. As these mountainous areas include steep slopes, and undeveloped land, a number of image tiles fell within zones that did not contain any buildings. After careful evaluation, 330 such images were excluded from annotation, as they lacked identifiable structures. These images were therefore not georeferenced, cropped, renamed, or labelled.

However, to support potential reconstruction of the broader landscape or for contextual understanding, the unannotated images have been retained in a separate folder within the dataset archive on Zenodo. The remaining 696 images—with visible buildings and flood-related damage—form the core of the annotated dataset.
}

\bigskip
\dssectionheader{Preprocessing/cleaning/labeling}

\dsquestionex{Was any preprocessing/cleaning/labeling of the data done (e.g., discretization or bucketing, tokenization, part-of-speech tagging, SIFT feature extraction, removal of instances, processing of missing values)?}{If so, please provide a description. If not, you may skip the remainder of the questions in this section.}

\dsanswer{No pre-processing of images was done. However, the JPEG images obtained through downloading from Google Earth Pro were converted to TIFF after geo-tagging and annotation with QGIS.
}

\dsquestionex{Was the “raw” data saved in addition to the preprocessed/cleaned/labeled data (e.g., to support unanticipated future uses)?}{If so, please provide a link or other access point to the “raw” data.}

\dsanswer{Yes, the raw images have been made available on Zenodo\cite{KuyeseraAI2024}.
}

\dsquestionex{Is the software used to preprocess/clean/label the instances available?}{If so, please provide a link or other access point.}

\dsanswer{The scripts to extract and preprocess the images can be made available upon reasonable request to the corresponding author.
}

\dsquestion{Any other comments?}

\dsanswer{No.}

\bigskip
\dssectionheader{Uses}

\dsquestionex{Has the dataset been used for any tasks already?}{If so, please provide a description.}

\dsanswer{The dataset won Round 1 of the  IRCAI, Zindi, and AWS’s AI for Equity Challenge and was used in Round 2 of the competition on Zindi for object detection and damage classification tasks \cite{zindiKAI}.
}

\dsquestionex{Is there a repository that links to any or all papers or systems that use the dataset?}{If so, please provide a link or other access point.}

\dsanswer{The dataset is accessible on Zenodo\cite{KuyeseraAI2024}
}
\dsquestion{What (other) tasks could the dataset be used for?}
\begin{enumerate}
    \item Environmental Visualization: Visualizing flood damage across landscapes.
    \item Urban Planning: Supporting decisions on relocation, the development of drainage systems, and other infrastructure improvements.
\end{enumerate}
}

\dsquestionex{Is there anything about the composition of the dataset or the way it was collected and preprocessed/cleaned/labeled that might impact future uses?}{For example, is there anything that a future user might need to know to avoid uses that could result in unfair treatment of individuals or groups (e.g., stereotyping, quality of service issues) or other undesirable harms (e.g., financial harms, legal risks) If so, please provide a description. Is there anything a future user could do to mitigate these undesirable harms?}

\dsanswer{No.
}

\dsquestionex{Are there tasks for which the dataset should not be used?}{If so, please provide a description.}

\dsanswer{The dataset can be used for any research where it is applicable.

}

\dsquestion{Any other comments?}

\dsanswer{No.
}

\bigskip
\dssectionheader{Distribution}

\dsquestionex{Will the dataset be distributed to third parties outside of the entity (e.g., company, institution, organization) on behalf of which the dataset was created?}{If so, please provide a description.}

\dsanswer{The dataset has been made publicly available.
}

\dsquestionex{How will the dataset will be distributed (e.g., tarball on website, API, GitHub)}{Does the dataset have a digital object identifier (DOI)?}

\dsanswer{The dataset has been distributed as a compressed (.zip) file containing the Images and corresponding json files on Zenodo\cite{KuyeseraAI2024}.
}

\dsquestion{When will the dataset be distributed?}

\dsanswer{It has been distributed on Zenodo
}

\dsquestionex{Will the dataset be distributed under a copyright or other intellectual property (IP) license, and/or under applicable terms of use (ToU)?}{If so, please describe this license and/or ToU, and provide a link or other access point to, or otherwise reproduce, any relevant licensing terms or ToU, as well as any fees associated with these restrictions.}

\dsanswer{Refer to Zenodo's \href{https://about.zenodo.org/terms/}{terms of use}
}

\dsquestionex{Have any third parties imposed IP-based or other restrictions on the data associated with the instances?}{If so, please describe these restrictions, and provide a link or other access point to, or otherwise reproduce, any relevant licensing terms, as well as any fees associated with these restrictions.}

\dsanswer{No.
}

\dsquestionex{Do any export controls or other regulatory restrictions apply to the dataset or to individual instances?}{If so, please describe these restrictions, and provide a link or other access point to, or otherwise reproduce, any supporting documentation.}

\dsanswer{No.
}

\dsquestion{Any other comments?}

\dsanswer{No.
}

\bigskip
\dssectionheader{Maintenance}

\dsquestion{Who will be supporting/hosting/maintaining the dataset?}

\dsanswer{\href{https://kailab.tech/}{Kuyesera AI Lab}
}

\dsquestion{How can the owner/curator/manager of the dataset be contacted (e.g., email address)?}

    \dsanswer{Contact the corresponding author
}

\dsquestionex{Is there an erratum?}{If so, please provide a link or other access point.}

\dsanswer{No.
}

\dsquestionex{Will the dataset be updated (e.g., to correct labeling errors, add new instances, delete instances)?}{If so, please describe how often, by whom, and how updates will be communicated to users (e.g., mailing list, GitHub)?}

\dsanswer{No
}

\dsquestionex{If the dataset relates to people, are there applicable limits on the retention of the data associated with the instances (e.g., were individuals in question told that their data would be retained for a fixed period of time and then deleted)?}{If so, please describe these limits and explain how they will be enforced.}

\dsanswer{No.
}

\dsquestionex{Will older versions of the dataset continue to be supported/hosted/maintained?}{If so, please describe how. If not, please describe how its obsolescence will be communicated to users.}

\dsanswer{No this is the only version of the dataset, look out for any communication regarding the dataset on the KAI website.
}

\dsquestionex{If others want to extend/augment/build on/contribute to the dataset, is there a mechanism for them to do so?}{If so, please provide a description. Will these contributions be validated/verified? If so, please describe how. If not, why not? Is there a process for communicating/distributing these contributions to other users? If so, please provide a description.}

\dsanswer{Contributions to the dataset can be submitted to the authors for verification.
}

\dsquestion{Any other comments?}

\dsanswer{No.
}

\bibliography{mwBTFreddy}
\end{multicols}
\end{singlespace}

\end{document}